\pdfoutput=1

\documentclass[11pt]{article}

\usepackage[]{EMNLP2023}

\usepackage{times}
\usepackage{latexsym}

\usepackage[T1]{fontenc}

\usepackage[utf8]{inputenc}

\usepackage{microtype}

\usepackage{inconsolata}
\usepackage{amssymb}
\usepackage{xspace}
\usepackage{hhline}
\usepackage{tabularx}
\usepackage{multirow}
\usepackage{adjustbox,booktabs}
\usepackage{mathtools}
\usepackage[multiple]{footmisc}
\usepackage{footnote}
\usepackage{CJKutf8}

\makesavenoteenv{tabular}
\makesavenoteenv{table}
\usepackage{makecell}
\usepackage{cleveref}
\crefformat{footnote}{#2\footnotemark[#1]#3}
\newcommand{\secref}[2][]{Section#1~\ref{#2}\xspace}

\newcommand{\tabref}[2][]{Table#1~\ref{#2}\xspace}
\newcommand{\eqnref}[2][]{Eqn#1.~(\ref{#2})\xspace}

\newcommand{\ex}[1]{\textit{#1}\xspace}

\title{Unsupervised Lexical Simplification with Context Augmentation}

\author{Takashi Wada$^{1,2}$ \qquad
Timothy Baldwin$^{1,2}$ \qquad
Jey Han Lau$^{1}$ \\
    $^1$ School of Computing and Information Systems, The University of Melbourne \\
    $^2$ Department of Natural Language Processing, MBZUAI \\[0ex]
    \texttt{twada@student.unimelb.edu.au} \qquad \texttt{tb@ldwin.net}\qquad \texttt{jeyhan.lau@gmail.com}
    }    

\begin{document}
\maketitle
\begin{abstract}
We propose a new unsupervised lexical simplification method that uses only monolingual data and pre-trained language models. Given a target word and its context, our method generates substitutes based on the target context and also additional contexts sampled from monolingual data. We conduct experiments in English, Portuguese, and Spanish on the TSAR-2022 shared task, and show that our model substantially outperforms other unsupervised systems across all languages. We also establish a new state-of-the-art by ensembling our model with GPT-3.5. Lastly, we evaluate our model on the SWORDS lexical substitution data set, achieving a state-of-the-art result.\footnote{Code is available at: \url{https://github.com/twadada/lexsub_decontextualised}.}
\end{abstract}

\section{Introduction}
Lexical simplification is the task of replacing a word in context with an easier term without changing its core meaning, to make text easier to read for non-technical audiences, non-native speakers, or people with cognitive disabilities (e.g.\ dyslexia).

One common approach \cite{li-etal-2022-mantis,qiang2020BERTLS,DBLP:journals/corr/abs-2006-14939} is to use a masked language model (MLM) such as BERT \cite{bert} and predict substitutes via word prediction over the masked target word. However, one limitation is that it critically relies on the target context being discriminative of the semantics of the target word, which is not always the case. Given this, we propose a new unsupervised method that performs context augmentation. Specifically, we sample sentences that contain the target word from monolingual data, and identify substitutes that can replace the word in the target and sampled sentences. Based on our experiments in English, Portuguese, and Spanish over the TSAR-2022 shared task \cite{saggion-etal-2022-findings}, we show that our model comfortably outperforms other unsupervised models. We also establish a new state-of-the-art by ensembling our model with InstructGPT \cite{gpt-3.5}, and further demonstrate the effectiveness of the method over the related task of lexical substitution.

\section{Method}
We propose a {\it fully} unsupervised model using pre-trained language models (without fine-tuning) and monolingual data. Given the target word $x$ and context $c_{x}$, our model generates substitutes for $x$   based on not only $c_{x}$ but also augmented contexts sampled from monolingual data.  

\subsection{Generation Based on the Target Context} \label{generation_tgt}
To generate substitutes of $x$ given $c_x$, we extend the lexical substitution approach of \citet{wada-etal-2022-unsupervised},\footnote{Lexical substitution is closely related to lexical simplification, with no constraint on the lexical complexity of $x$.} which generates an aptness score $S(y|x, c_x)$ for each word $y\in V$ as follows:\footnote{We set $|V|$ to 20,000 and 30,000 for the lexical simplification and substitution tasks, respectively.}
\begin{equation}
{S}(y|x,c_t)=\underset{k}{\mathrm{max~}}{\mathrm{cos}(f^{k}(y),f(x, c_x))},\footnote{We also add the \ex{global similarity} term to \eqnref[]{eqn_coling} as proposed by \citet{wada-etal-2022-unsupervised}.}\label{eqn_coling}
\end{equation}
where $\mathrm{cos}$ denotes cosine similarity; $f(x, c_x)$ denotes the contextualised embedding of $x$ in $c_x$;\footnote{Obtained by feeding $c_x$ into a pre-trained LM~$f$ and averaging the embeddings of $x$ across multiple layers.} and $f^{k}(y)$ denotes the {\it decontextualised} embeddings of $y$, represented by $K$-clustered embeddings: $f^{1}(y), ...f^{K}(y)$, which are obtained by first sampling 300 sentences that contain $y$ from monolingual corpora, and clustering the contextualised embeddings of $y$ using $K$-means ($K=4$). For each cluster $k$, $f^{k}(y)$ is calculated as $\dfrac{1}{|C_{y,k}|}\sum_{ c'_y \in C_{y,k}}{f(y,{c'_y})}$,
where $C_{y,k}$ denotes sentences that contain $y$ and belong to the cluster $k$. 

While this is the state-of-the-art unsupervised method on the SWORDS lexical substitution data set \cite{swords}, one major limitation is that ${S}(y|x,c_x)$ in \eqnref[]{eqn_coling} heavily depends on $f(x, c_x)$, suggesting that if the meaning of $x$ is not well captured by $f(x, c_x)$, it may retrieve erroneous substitutes. In fact, this is often the case in lexical {\it simplification}, where $x$ is usually a rare word and gets segmented into subword tokens (in which case $f(x, c_x)$ is represented by the average of the subword embeddings). For instance, given the target word \ex{bole}, the model retrieves \ex{toe}  as one of the top-10 substitutes, likely because the segmented \ex{bol~{\#\#e}} and \ex{to~{\#\#e}} share the same subword \ex{{\#\#e}}, suggesting that words that share the same token(s) tend to have similar representations regardless of their semantic similarity. To mitigate this, when $x$ is tokenised into multiple tokens, we
add the term {$\alpha \mathrm{cos}(E(x),E(y))$} to \eqnref[]{eqn_coling}, where $\alpha$ is a scalar value and $E(x)$ and $E(y)$ are pre-trained static embeddings of $x$ and $y$; we use fastText \cite{fasttext} for this purpose. Since static embeddings are pre-trained with a large vocabulary size (e.g.\ 200k words), they tend to represent the semantics of rare words better than averaging embeddings of their (suboptimally tokenised) subwords.\footnote{While fastText similarly makes use of character $n$-grams to represent words, it also trains a unique representation for each word, which is not shared with any other words (e.g.\ the word embedding of \ex{her} is constructed by its character $n$-grams plus the special sequence \ex{<her>}, where \ex{<} and \ex{>} correspond to the beginning and end of token). We also tried using GloVe \cite{glove} instead of fastText in English, and observed comparable results.}

We tune $\alpha$ on the dev set and set it to 0.2, 0.7 and 0.6 for English, Spanish, and Portuguese, respectively.  For embedding model $f$, we use DeBERTa-V3 \cite{deberta-v3} for English, and monolingual BERT models for Spanish and Portuguese  \cite{CaneteCFP2020,souza2020bertimbau}. We extract the $M_1=15$ words with the highest scores.

\subsection{Generation with Context Augmentation} \label{generation_sample}

Following previous work on lexical simplification \cite{li-etal-2022-mantis,qiang2020BERTLS,DBLP:journals/corr/abs-2006-14939}, we also generate substitutes based on MLM prediction, by replacing $x$ with a mask token and performing word prediction. In this approach, the predictions are not affected by the embedding quality or tokenisation of $x$. However, if we rely solely on the target context $c_x$ as in previous work, the model has difficulty predicting substitutes when the context is not very specific; e.g.\ \ex{The \underline{bole} was cut into pieces}.\footnote{While previous work concatenates the masked sentence with the original (unmasked) sentence, it does not completely solve this problem.} To address this problem, we perform context augmentation using monolingual data. Following the process of generating decontextualised embeddings in \citet{wada-etal-2022-unsupervised}, we sample 300 sentences that contain $x$ from monolingual corpora and cluster them using $K$-means ($K=4$).\footnote{Note that \citet{wada-etal-2022-unsupervised} sample sentences for generating $f^{k}(y)$, not for augmenting the contexts of $x$.} For each sentence in cluster $k$, we replace $x$ with a mask token and feed it into T5 \cite{t5} in English, or mT5 \cite{mt5} in Spanish and Portuguese, to generate 20 substitutes using beam search,\footnote{In English, when the mask token directly follows the article \ex{an} or \ex{a}, we replace one of them with the other and feed the modified sentence to T5 and generate another 20 outputs. This way, we can mitigate the morphophonetic bias reported in \citet{wada-etal-2022-unsupervised} (e.g.\ most of the generated substitutes for \ex{accord} start with a vowel sound).}  and retain those that contain only one word (which can comprise multiple subwords). Then, within each cluster $k$, we aggregate the substitutes across all sentences $c'_x \in C_{x,k}$ and extract the $M_2=25$ most-generated words. For each substitute candidate $y$, we calculate the score ${\tilde{S}}(y|x, c_x)$ as:
\begin{gather}
\tilde{S}(y|x, c_x)=\sum_{k}w_{k} \sum_{c'_x \in C_{x,k}}{\mathrm{I}(y|c'_x))},\label{t5_score}
\end{gather}
where ${\mathrm{I}(y|c'_x))}$ denotes a function that returns 1 if $y$ is generated by T5 given the context $c'_x$, and 0 otherwise; and $w_k$ denotes the number of substitutes in the cluster $k$ that overlap with the $M_1$ words generated from the target context $c_x$ in \secref[]{generation_tgt}.\footnote{In English, $\max_{k} w_k$ ranges from 0 to 12 (5.9 on average) with $M_1 = 15$ and $M_2 = 25$.}
Here,  $w_k$ roughly corresponds to the semantic relevance of the cluster~$k$ to $c_x$; e.g.\ if $w_k=0$, the candidates in the cluster $k$ would reflect a different sense of $x$ from the one in the target context and hence is not considered.\footnote{When $w_k =0$ for all clusters, we set $w_k$ to 1.} Intuitively, this scoring function favours substitutes that appear frequently in sampled contexts, weighted by how semantically relevant the substitute's cluster is to the original context --- we will show its effectiveness with an example in \secref{ablation_study}. 
Finally, we retrieve the $M_2$ words with the highest scores and combine them with the $M_1$ candidates generated from $c_x$.

\subsection{Reranking}
Given $M_1{+}M_2$ candidates\footnote{Following \citet{wada-etal-2022-unsupervised}, we discard lexically-similar candidates, as measured by edit distance.} (potentially with overlap), we rerank them using four different metrics: (i) embedding similarity; (ii) LM perplexity; (iii) word frequency; and (iv) $\tilde{S}(y|x,c_x)$ in \eqnref[]{t5_score}. For the first metric, we use the reranking method proposed by \citet{wada-etal-2022-unsupervised}. For each candidate $y$, they replace $x$ in $c_x$ with $y$ and calculate the cosine similarity between the contextualised embeddings $f(x, c_x)$ and $f(y, c_x)$.\footnote{As in \secref[]{generation_tgt}, we add $\alpha \mathrm{cos}(E(x),E(y))$ when $x$ is segmented into multiple tokens.} For the LM perplexity metric, we replace $x$ in $c_x$ with a mask token and calculate the probability of generating $y$ using T5; this score helps measure the syntactic fit of $y$ in $c_x$. The third metric corresponds to the frequency of $y$ in monolingual data,\footnote{We use the {\it wordfreq} Python library \cite{robyn_speer_2022_7199437}.} which serves as a proxy for lexical simplicity. Finally, the last metric measures how often $y$ can substitute $x$ in the augmented contexts. Using each metric, we obtain four independent rankings $R_1, R_2, R_3, R_4$ and calculate their weighted sum: $r_1R_1{+}r_2R_2{+}r_3R_3{+}r_4R_4$, 
 which is then sorted in ascending order to produce the final ranking. We tune the weights $\{r_1, r_2, r_3, r_4\}$ based on the dev set for each language; $\{5, 1, 1, 1\}$, $\{3, 1, 0, 3\}$, and $\{3, 1, 0, 2\}$ for English, Spanish and Portuguese, respectively.

\section{Experiments}
\begin{table*}[t!]
\begin{center}
\begin{adjustbox}{max width=\textwidth}
\begin{tabular}{lllllllllllllllllllll}
\toprule
\multirow{2}{*}{Model}
&\multicolumn{3}{c}{\textbf{ACC@1}}
&&\multicolumn{3}{c}{\textbf{ACC@3@Top1}}
&&\multicolumn{3}{c}{\textbf{MAP@3}}
&&\multicolumn{3}{c}{{\textbf{Potential@3}}}\\
  \cmidrule{2-4}
  \cmidrule{6-8}
  \cmidrule{10-12}
  \cmidrule{14-16}

&\multicolumn{1}{c}{en}&\multicolumn{1}{c}{pt}&\multicolumn{1}{c}{es}
&&\multicolumn{1}{c}{en}&\multicolumn{1}{c}{pt}&\multicolumn{1}{c}{es}
&&\multicolumn{1}{c}{en}&\multicolumn{1}{c}{pt}&\multicolumn{1}{c}{es}
&&\multicolumn{1}{c}{en}&\multicolumn{1}{c}{pt}&\multicolumn{1}{c}{es}\\\midrule

\multicolumn{16}{c}{{InstructGPT (UniHD)}}\\
  \midrule
  GPT-3.5-text-davinci-002-zero
&77.2&63.6&57.1&&57.1&51.6&45.1&&50.9&41.1&35.3&&89.0&78.6&69.0\\

GPT-3.5-text-davinci-002-ens
&81.0&77.0&65.2&&68.6&62.3&57.9&&58.3&50.1&42.8&&96.2&91.7&82.1\\
GPT-3.5-turbo-zero
&82.8&79.4&64.4&&68.1&63.6&50.5&&60.9&51.1&45.2&&92.8&88.8&75.0\\
GPT-3.5-turbo-ens
&\underline{87.4}&\underline{85.8}&\underline{76.4}&&\underline{71.8}&\underline{73.5}&\underline{62.2}&&\underline{65.5}&\underline{58.7}&\underline{55.9}&&\underline{97.3}&\textbf{\underline{97.3}}&\underline{89.1}
\\

GPT-3.5-turbo-ens (w/o context)
&82.6&84.5&76.1&&70.8&69.8&59.5&&64.8&57.3&54.1&&94.6&94.9&88.3\\

\midrule
\multicolumn{16}{c}{{Unsupervised}}\\\midrule
MANTIS/GMU-WLV/PresiUniv 
&65.7&48.1&37.0&&53.9&39.6&32.9&&47.3&28.2&21.5&&87.7&68.7&58.4\\
LSBert 
&59.8&32.6&28.8&&53.1&28.6&18.2&&40.8&19.0&18.7&&82.3&49.5&49.5\\
\citet{wada-etal-2022-unsupervised} 
&64.1&39.3&21.7&&50.7&32.9&14.7&&43.3&24.1&13.0&&86.6&59.4&31.0
\\

\citet{wada-etal-2022-unsupervised} + fastText
&64.6&51.9&32.3&&51.2&42.8&25.0&&43.7&31.0&19.5&&86.9&70.9&49.2\\

\textbf{OURS}
&\underline{79.9}&\underline{61.5}&\underline{47.8}&&\underline{63.5}&\underline{52.7}&\underline{37.0}&&\underline{57.5}&\underline{38.0}&\underline{30.0}&&\underline{94.1}&\underline{83.2}&\underline{71.5}

\\\midrule\midrule

GPT-3.5-turbo-ens + WordFreq
&\textbf{89.3}&85.6&\textbf{78.3}&&73.2&74.9&65.2&&68.2&59.9&57.6&&97.9&\textbf{97.3}&89.4
\\

+ \textbf{OURS}
&\textbf{89.3}&\textbf{86.4}&77.7&&\textbf{75.1}&\textbf{76.7}&\textbf{66.8}&&\textbf{69.9}&\textbf{61.1}&\textbf{59.1}&&\textbf{98.7}&\textbf{97.3}&\textbf{89.9}\\

\bottomrule
\end{tabular}
\end{adjustbox}
\end{center}
\caption{The results on lexical simplification. ``-zero/ens'' denote the zero-shot/ensemble models, and ``w/o context'' indicates the performance without access to the target context. The best scores among InstructGPT and unsupervised models are underlined, and the overall best scores are boldfaced.}
\label{result_tsar}
\end{table*}
 
\subsection{Data and Evaluation}
We experiment on the TSAR-2022 shared task on multilingual lexical simplification \cite{saggion-etal-2022-findings,10.3389/frai.2022.991242,ferres-saggion-2022-alexsis,north-etal-2022-alexsis}. We use its trial data as our dev set (about 10 instances per language) and evaluate models on the test set, which contains about 370 instances per language. Evaluation is according to four metrics: \textbf{Accuracy@1} = \% of instances for which the top-1 substitute matches one of the gold candidates; \textbf{Accuracy@k@top1} = \% of instances where one of the top-$k$ substitutes matches the top-1 gold label; \textbf{Potential@k} = \% of instances where at least one of the top-$k$ substitutes is included in the gold candidates; and \textbf{MAP@k} = the mean average precision of the top-$k$ candidates. 

\subsection{Baselines}

We compare our method against several systems submitted to the shared task. In all languages, \textbf{UniHD} \cite{aumiller-gertz-2022-unihd} is by far the best system across all metrics. It prompts GPT-3.5 (\ex{text-davinici-002}, a.k.a.\ InstructGPT: \citet{gpt-3,gpt-3.5}) to provide ten easier alternatives for the target word $x$ given $c_x$, in two variants: \textbf{zero-shot} and \textbf{ensemble}. The former generates substitutes based on the target word and context only, whereas the latter ensembles the predictions with six different prompts and temperatures; among them, four prompts include one or two question--answer pairs retrieved from the dev set to allow InstructGPT to perform in-context learning (as detailed in \tabref[]{prompt_template} in Appendix). While the ensemble model achieves the best results across all languages, it is not exactly comparable with the other systems as InstructGPT is {\it supervised} on various tasks with human feedback. As such, we also include the second-best systems (which differ for each language) as baselines, namely: \textbf{MANTIS} \cite{li-etal-2022-mantis}, \textbf{GMU-WLV} \cite{north-etal-2022-gmu}, and \textbf{PresiUniv} \cite{whistely-etal-2022-presiuniv}. We also include the shared task baseline \textbf{LSBert} \cite{qiang2020BERTLS,DBLP:journals/corr/abs-2006-14939}. All of these systems are based on pre-trained MLMs like BERT \cite{bert} and RoBERTa \cite{roberta}, and three of them also employ static word embeddings similarly to our model. Lastly, we also include \citet{wada-etal-2022-unsupervised} with and without fastText in our baselines.

\subsection{Results}
\tabref[]{result_tsar} presents the results in English, Portuguese, and Spanish. The first five rows are based on InstructGPT: the first two show the zero-shot/ensemble performance reported in \citet{aumiller-gertz-2022-unihd}, and the next two show the results when we replace {\it text-davinci-002} with {\it gpt-3.5-turbo}. The results show that {\it gpt-3.5-turbo} substantially outperforms {\it text-davinci-002}. The last row for InstructGPT shows the result when we prompt the model to provide simplified alternatives for $x$ \ex{without the target context} (shown as \textbf{``w/o context''}), which indicates that the model performs very well {even without access to the target context}. This result demonstrates that the model has memorised lists of synonyms, and that most instances are not very context-dependent; we will return to discuss this in Appendix \ref{error_analysis}.

The next five rows (marked ``Unsupervised'') show the performance of the unsupervised models, including ours. Our model clearly outperforms the other systems across all languages. In English, it even outperforms the zero-shot GPT-3.5-turbo in Potential@3 (94.1 vs.\ 92.8) despite the substantial differences between these models in terms of the model size (i.e.\ 435M and 800M parameters are used for DeBERTa-V3 and T5, respectively, and 175B parameters for GPT-3.5) and the language resources to use (i.e.\ our model employs monolingual data only while GPT-3.5 is instructed with human feedback). The strong performance in English is largely owing to the use of better LMs (DeBERTa-V3 and T5) compared to the ones used in Spanish and Portuguese (BERT and mT5), as evidenced by the substantial performance drop when we use BERT and mT5 for English.\footnote{The exact scores are 69.4, 58.7, 45.0, and 88.2 for ACC@1, ACC@3@Top1, MAP@3, and Potential@3, resp.} The comparison of \citet{wada-etal-2022-unsupervised} with and without fastText demonstrates the effectiveness of including static embeddings, especially in Portuguese and Spanish. This is because the vocabulary size of Portuguese/Spanish BERT is much smaller than that of DeBERTa-V3 (30/31k vs.\ 128k), and a large number of target words are segmented into subwords and embedded poorly. Lastly, we try ensembling: (1) the six rankings from GPT-3.5-turbo-ens; (2) the word frequency ranking (which we find boosts performance); and (3) the final ranking of OURS. The last two rows show the performance for (1) + (2) vs.\ (1) + (2) + (3). The ensemble of eight rankings including our method establishes a new state-of-the-art across all languages in most metrics, suggesting that our model is somewhat complementary to InstructGPT. 
In Appendix, we provide more detailed results (\tabref[]{result_tsar_all}) and error analysis (Appendix \ref{error_analysis}).

\begin{table}[t]
\begin{center}
\begin{adjustbox}{max width=\columnwidth}
\begin{tabular}{ccc@{\;}ccc@{\;}cc@{\;}}
\toprule
\multirow{2}{*}{Model} &\multicolumn{2}{c}{Lenient}&&\multicolumn{2}{c}{{Strict}}\\
  \cmidrule{2-3}
  \cmidrule{5-6}
  &$F_a$&$F_c$ && $F_a$&$F_c$\\\midrule
\multirow{1}{*}{GPT-3-davinci} &\textbf{34.6}& 49.0 &&22.7 &{36.3} \\
\multirow{1}{*}{GPT-3.5-turbo-ens} 
&32.5&67.5&&27.2&51.2 \\
\multirow{1}{*}{\citet{wada-etal-2022-unsupervised}} &33.6&{65.8}&&{24.5}&{39.9}\\

\multirow{1}{*}{\citet{qiang-etal-2023-parals}} &--&--&&24.9&40.1\\
\multirow{1}{*}{\textbf{OURS}} &33.3&66.2&&25.3&41.8\\
\multirow{1}{*}{\textbf{OURS} + GPT-3.5-turbo-ens} 
&33.1&\textbf{69.8}&&\textbf{28.2}&\textbf{52.2}\\
\bottomrule
\end{tabular}
\end{adjustbox}
\end{center}
\caption{The results on lexical substitution.}
\label{result_swords}
\end{table}

\begin{table*}[t!]
\begin{center}
\begin{adjustbox}{max width=\textwidth}
\begin{tabular}{llllllllllllllllllllllllll}
\toprule
\multirow{2}{*}{Method}
&\multirow{2}{*}{\textbf{ACC@1}}
&&\multicolumn{3}{c}{\textbf{ACC@k@Top1}}
&&\multicolumn{3}{c}{\textbf{MAP@k}}
&&\multicolumn{3}{c}{{\textbf{Potential@k}}}\\
  \cmidrule{4-6}
  \cmidrule{8-10}
  \cmidrule{12-14}
&
&&\multicolumn{1}{c}{\textbf{k=1}}&\multicolumn{1}{c}{\textbf{k=2}}&\multicolumn{1}{c}{\textbf{k=3}}
&&\multicolumn{1}{c}{\textbf{k=3}}&\multicolumn{1}{c}{\textbf{k=5}}&\multicolumn{1}{c}{\textbf{k=10}}
&&\multicolumn{1}{c}{\textbf{k=3}}&\multicolumn{1}{c}{\textbf{k=5}}&\multicolumn{1}{c}{\textbf{k=10}}
\\

  \midrule
Soft Retrieval&\textbf{63.1}
&&\textbf{32.4}&\textbf{45.0}&\textbf{51.1}
&&\textbf{41.8}&\textbf{30.9}&\textbf{18.8}
&&\textbf{82.9}&\textbf{89.2}&93.3\\
Hard Retrieval&60.5
&&30.7&43.3&50.9
&&40.3&29.7&18.1
&&82.4&86.6&91.9\\
No Clustering&62.4
&&31.6&44.7&{51.0}
&&41.3&30.8&\textbf{18.8}
&&82.4&88.9&\textbf{94.2}\\

\bottomrule
\end{tabular}
\end{adjustbox}
\caption{The results on the lexical simplification task using different cluster-retrieval methods in \eqnref[]{t5_score}. The scores are averaged over English, Portuguese, and Spanish. ``Soft Retrieval'' indicates our original method proposed in \eqnref[]{t5_score}, and ``Hard Retrieval'' denotes when we set $w_k = 1$ for the closest cluster and $w_k = 0$ otherwise. The last row indicates when we set $w_k = 1$ for all clusters, which is equivalent to performing no clustering.}\label{ablation_table}
\end{center}
\end{table*}

\begin{table}[t!]
\begin{center}
\begin{adjustbox}{max width=\columnwidth}
\begin{tabular}{m{0.38\linewidth}m{0.79\linewidth}}
\toprule
\multicolumn{1}{l}{ Context ($x$ = \ex{elite}) }&Syria is overwhelmingly Sunni, but President Bashar Assad and the ruling \underline{elite} belong to the minatory Alawite sect.\\\midrule
\multicolumn{1}{l}{\citet{wada-etal-2022-unsupervised}} &\textbf{establishment}, hierarchy, wealthy
\\\midrule
Cluster1 ($w_k=0$) &special, military, small
\\
Cluster2 ($w_k=5$)&\textbf{class}, political, \textbf{privileged}
\\
Cluster3 ($w_k=1$)&exclusive, international, prestigious
\\
Cluster4 ($w_k=0$)&top, professional, great
\\
\cmidrule{1-2}
\multirow{1}{*}{Soft Retrieval} &\textbf{class}, \textbf{privileged}, political
\\
\multirow{1}{*}{No Clustering} &top, professional, exclusive
\\\midrule
\textbf{OURS} &\textbf{class}, \textbf{establishment}, leadership
\\

\bottomrule
\end{tabular}
\end{adjustbox}
\end{center}
\caption{Top-3 substitutes generated based on the target context (\citet{wada-etal-2022-unsupervised}) and the augmented contexts (\secref{generation_sample}). The values for $w_k$ denote the weights for each cluster in \eqnref{t5_score}. ``OURS'' reranks the candidates of \citet{wada-etal-2022-unsupervised} and ``Soft Retrieval''. The words included in the gold labels are boldfaced.}\label{ablation_example}
\end{table}

\subsection{Experiment on Lexical Substitution}

We also evaluate our model on the English lexical {\it substitution} task over the SWORDS data set \cite{swords}. For lexical substitution, there is no restriction on lexical simplicity, so we drop the word frequency feature in reranking (i.e.\ set $r_3$ to $0$).\footnote{We also double $M1/M2$ to $30/50$ to make sure that our model provides top-50 words, following \citet{wada-etal-2022-unsupervised}.} \tabref[]{result_swords} shows the results in the {\it lenient} and  {\it strict} settings.\footnote{In the lenient setting, generated words are filtered out if their aptness scores are not annotated in SWORDS, whereas in the strict setting, all words are considered in the evaluation.} $F_a$ and $F_c$ denote the F1 scores given two different sets of gold labels $a$ and $c$, where $a \subset c$. In the strict setting, our model outperforms the best unsupervised model of \citet{wada-etal-2022-unsupervised} and also the (non-LLM) state-of-the-art \ex{semi-supervised} model of \citet{qiang-etal-2023-parals}, which employs BLEURT \cite{sellam-etal-2020-bleurt} and a sentence-paraphrasing model, both of which are pre-trained on labelled data. Lastly, we also ensemble our model with the six outputs of GPT-3.5, and establish a new state-of-the-art.

\section{Ablation Study}\label{ablation_study}

We perform ablation studies on the effect of clustering in \eqnref[]{t5_score}, and present the results in \tabref[]{ablation_table}; the scores are averaged over English, Portuguese, and Spanish. ``Soft Retrieval'' indicates the performance when we take the weighted sum of the clusters as we propose in \eqnref[]{t5_score}; ``Hard Retrieval'' denotes when we set $w_k = 1$ for the closest cluster and $w_k = 0$ otherwise; and ``No Clustering'' denotes when we set $w_k = 1$ for all the clusters, which is equivalent to performing no clustering. The table shows that our proposed method performs the best overall, albeit with a small margin over ``No Clustering''. In fact, this is more or less expected since the majority of target words are used in their predominant senses (as evidenced by the strong performance of GPT-3.5 w/o context in \tabref[]{result_tsar}), in which case, retrieving the most-generated words across all sampled sentences would suffice to produce good substitutes.

\tabref[]{ablation_example} shows one example where clustering plays a crucial role (more predictions plus another example are shown in \tabref{examples} in Appendix \ref{ablation_study_appendix}). In this example, the target word \ex{elite} is used as a noun meaning ``a select group'', but all clusters except for Cluster 2 produce the substitutes for \ex{elite} in adjectival senses. Therefore, if we naïvely aggregate the words across all clusters (``No Clustering''), we end up retrieving adjectives such as \ex{top} and \ex{professional}, whereas our weighted-sum approach (``Soft Retrieval'') successfully extracts good substitutes from the relevant clusters. 

\section{Related Work}
Recent lexical simplification models are based on generating substitute candidates using MLM prediction and reranking, using features such as fastText embedding similarities and word frequency \cite{qiang2020BERTLS,li-etal-2022-mantis}. Some also use external tools or resources such as POS taggers or paraphrase databases \cite{DBLP:journals/corr/abs-2006-14939,whistely-etal-2022-presiuniv}. However, \citet{aumiller-gertz-2022-unihd} show that GPT-3.5 substantially outperforms previous models on the TSAR-2022 shared task. Similar to this work, our prior work \cite{wada-etal-2023-unsupervised} samples sentences from monolingual corpora and use them to paraphrase multiword expressions with literal expressions (composed of 1 or 2 words).

\section{Conclusion}
We propose a new unsupervised lexical simplification method with context augmentation. We show that our model outperforms previous unsupervised methods, and by combining our model with InstructGPT, we achieve a new state-of-the-art for lexical simplification and substitution.

\section{Limitations}\label{sec_limitations}
One limitation of our model is that it performs context augmentation using monolingual data, which incurs additional time and computational cost. However, if we construct a comprehensive list of complex words $X$ and sample sentences containing $x \in X$ in advance, we can pre-compute the generation counts: $\sum_{c'_x \in C_{x,k}}{\mathrm{I}(y|c'_x)}$ in \eqnref{t5_score} without considering the target context $c_x$ (which is required to calculate $w_k$ only). Therefore, we can still generate substitutes in an online manner during inference as long as the target word $x$ is included in $X$. 

Compared to the InstructGPT baseline, our model critically relies on word embeddings and MLM prediction, both of which hinge on word co-occurrence statistics. This sometimes results in wrongly predicting antonyms of the target word as substitutes due to the similarity of their surrounding contexts (e.g.\ \ex{famed} for \ex{infamous}; more specific examples and error types are shown in Appendix \ref{error_analysis}). On the other hand, InstructGPT benefits from supervision with human feedback and also makes use of lexical knowledge provided in various forms of texts during pre-training, including dictionaries, thesauri, and web discussions about meanings of words.\footnote{This is evidenced by the fact that InstuctGPT can answer some questions that ask for very specific knowledge of words and phrases, such as their etymology.} This is clearly one of the reasons why InstuctGPT substantially outperforms the other unsupervised systems, including ours; in fact, we find that it performs extremely well even {\it without access to the target context} (\tabref[]{result_tsar}), motivating a call for including more context-sensitive instances in lexical substitution/simplification data sets; more discussions follow in Appendix \ref{error_analysis}.

\bibliography{anthology}

\begin{thebibliography}{27}
\expandafter\ifx\csname natexlab\endcsname\relax\def\natexlab#1{#1}\fi

\bibitem[{Aumiller and Gertz(2022)}]{aumiller-gertz-2022-unihd}
Dennis Aumiller and Michael Gertz. 2022.
\newblock \href {https://aclanthology.org/2022.tsar-1.28} {{U}ni{HD} at
  {TSAR}-2022 shared task: Is compute all we need for lexical simplification?}
\newblock In \emph{Proceedings of the Workshop on Text Simplification,
  Accessibility, and Readability (TSAR-2022)}, pages 251--258, Abu Dhabi,
  United Arab Emirates (Virtual). Association for Computational Linguistics.

\bibitem[{Bojanowski et~al.(2017)Bojanowski, Grave, Joulin, and
  Mikolov}]{fasttext}
Piotr Bojanowski, Edouard Grave, Armand Joulin, and Tomas Mikolov. 2017.
\newblock \href {https://doi.org/10.1162/tacl_a_00051} {Enriching word vectors
  with subword information}.
\newblock \emph{Transactions of the Association for Computational Linguistics},
  5:135--146.

\bibitem[{Brown et~al.(2020)Brown, Mann, Ryder, Subbiah, Kaplan, Dhariwal,
  Neelakantan, Shyam, Sastry, Askell, Agarwal, Herbert-Voss, Krueger, Henighan,
  Child, Ramesh, Ziegler, Wu, Winter, Hesse, Chen, Sigler, Litwin, Gray, Chess,
  Clark, Berner, McCandlish, Radford, Sutskever, and Amodei}]{gpt-3}
Tom Brown, Benjamin Mann, Nick Ryder, Melanie Subbiah, Jared~D Kaplan, Prafulla
  Dhariwal, Arvind Neelakantan, Pranav Shyam, Girish Sastry, Amanda Askell,
  Sandhini Agarwal, Ariel Herbert-Voss, Gretchen Krueger, Tom Henighan, Rewon
  Child, Aditya Ramesh, Daniel Ziegler, Jeffrey Wu, Clemens Winter, Chris
  Hesse, Mark Chen, Eric Sigler, Mateusz Litwin, Scott Gray, Benjamin Chess,
  Jack Clark, Christopher Berner, Sam McCandlish, Alec Radford, Ilya Sutskever,
  and Dario Amodei. 2020.
\newblock \href
  {https://proceedings.neurips.cc/paper/2020/file/1457c0d6bfcb4967418bfb8ac142f64a-Paper.pdf}
  {Language models are few-shot learners}.
\newblock In \emph{Advances in Neural Information Processing Systems},
  volume~33, pages 1877--1901. Curran Associates, Inc.

\bibitem[{Cañete et~al.(2020)Cañete, Chaperon, Fuentes, Ho, Kang, and
  Pérez}]{CaneteCFP2020}
José Cañete, Gabriel Chaperon, Rodrigo Fuentes, Jou-Hui Ho, Hojin Kang, and
  Jorge Pérez. 2020.
\newblock Spanish pre-trained {BERT} model and evaluation data.
\newblock In \emph{PML4DC at ICLR 2020}.

\bibitem[{Devlin et~al.(2019)Devlin, Chang, Lee, and Toutanova}]{bert}
Jacob Devlin, Ming-Wei Chang, Kenton Lee, and Kristina Toutanova. 2019.
\newblock \href {https://doi.org/10.18653/v1/N19-1423} {{BERT}: Pre-training of
  deep bidirectional transformers for language understanding}.
\newblock In \emph{Proceedings of the 2019 Conference of the North {A}merican
  Chapter of the Association for Computational Linguistics: Human Language
  Technologies, Volume 1 (Long and Short Papers)}, pages 4171--4186,
  Minneapolis, Minnesota. Association for Computational Linguistics.

\bibitem[{Ferr{\'e}s and Saggion(2022)}]{ferres-saggion-2022-alexsis}
Daniel Ferr{\'e}s and Horacio Saggion. 2022.
\newblock \href {https://aclanthology.org/2022.lrec-1.383} {{ALEXSIS}: A
  dataset for lexical simplification in {S}panish}.
\newblock In \emph{Proceedings of the Thirteenth Language Resources and
  Evaluation Conference}, pages 3582--3594, Marseille, France. European
  Language Resources Association.

\bibitem[{He et~al.(2023)He, Gao, and Chen}]{deberta-v3}
Pengcheng He, Jianfeng Gao, and Weizhu Chen. 2023.
\newblock \href {https://openreview.net/forum?id=sE7-XhLxHA} {De{BERT}av3:
  Improving de{BERT}a using {ELECTRA}-style pre-training with
  gradient-disentangled embedding sharing}.
\newblock In \emph{The Eleventh International Conference on Learning
  Representations}.

\bibitem[{Lee et~al.(2021)Lee, Donahue, Jia, Iyabor, and Liang}]{swords}
Mina Lee, Chris Donahue, Robin Jia, Alexander Iyabor, and Percy Liang. 2021.
\newblock \href {https://doi.org/10.18653/v1/2021.naacl-main.345} {{SWORDS}: A
  benchmark for lexical substitution with improved data coverage and quality}.
\newblock In \emph{Proceedings of the 2021 Conference of the North American
  Chapter of the Association for Computational Linguistics: Human Language
  Technologies}, pages 4362--4379, Online. Association for Computational
  Linguistics.

\bibitem[{Li et~al.(2022)Li, Wiechmann, Qiao, and Kerz}]{li-etal-2022-mantis}
Xiaofei Li, Daniel Wiechmann, Yu~Qiao, and Elma Kerz. 2022.
\newblock \href {https://aclanthology.org/2022.tsar-1.27} {{MANTIS} at
  {TSAR}-2022 shared task: Improved unsupervised lexical simplification with
  pretrained encoders}.
\newblock In \emph{Proceedings of the Workshop on Text Simplification,
  Accessibility, and Readability (TSAR-2022)}, pages 243--250, Abu Dhabi,
  United Arab Emirates (Virtual). Association for Computational Linguistics.

\bibitem[{Liu et~al.(2019)Liu, Ott, Goyal, Du, Joshi, Chen, Levy, Lewis,
  Zettlemoyer, and Stoyanov}]{roberta}
Yinhan Liu, Myle Ott, Naman Goyal, Jingfei Du, Mandar Joshi, Danqi Chen, Omer
  Levy, Mike Lewis, Luke Zettlemoyer, and Veselin Stoyanov. 2019.
\newblock \href {http://arxiv.org/abs/1907.11692} {{RoBERTa}: {A} robustly
  optimized {BERT} pretraining approach}.
\newblock \emph{CoRR}, abs/1907.11692.

\bibitem[{North et~al.(2022{\natexlab{a}})North, Dmonte, Ranasinghe, and
  Zampieri}]{north-etal-2022-gmu}
Kai North, Alphaeus Dmonte, Tharindu Ranasinghe, and Marcos Zampieri.
  2022{\natexlab{a}}.
\newblock \href {https://aclanthology.org/2022.tsar-1.30} {{GMU}-{WLV} at
  {TSAR}-2022 shared task: Evaluating lexical simplification models}.
\newblock In \emph{Proceedings of the Workshop on Text Simplification,
  Accessibility, and Readability (TSAR-2022)}, pages 264--270, Abu Dhabi,
  United Arab Emirates (Virtual). Association for Computational Linguistics.

\bibitem[{North et~al.(2022{\natexlab{b}})North, Zampieri, and
  Ranasinghe}]{north-etal-2022-alexsis}
Kai North, Marcos Zampieri, and Tharindu Ranasinghe. 2022{\natexlab{b}}.
\newblock \href {https://aclanthology.org/2022.coling-1.529} {{ALEXSIS}-{PT}: A
  new resource for {P}ortuguese lexical simplification}.
\newblock In \emph{Proceedings of the 29th International Conference on
  Computational Linguistics}, pages 6057--6062, Gyeongju, Republic of Korea.
  International Committee on Computational Linguistics.

\bibitem[{Ouyang et~al.(2022)Ouyang, Wu, Jiang, Almeida, Wainwright, Mishkin,
  Zhang, Agarwal, Slama, Gray, Schulman, Hilton, Kelton, Miller, Simens,
  Askell, Welinder, Christiano, Leike, and Lowe}]{gpt-3.5}
Long Ouyang, Jeffrey Wu, Xu~Jiang, Diogo Almeida, Carroll Wainwright, Pamela
  Mishkin, Chong Zhang, Sandhini Agarwal, Katarina Slama, Alex Gray, John
  Schulman, Jacob Hilton, Fraser Kelton, Luke Miller, Maddie Simens, Amanda
  Askell, Peter Welinder, Paul Christiano, Jan Leike, and Ryan Lowe. 2022.
\newblock \href {https://openreview.net/forum?id=TG8KACxEON} {Training language
  models to follow instructions with human feedback}.
\newblock In \emph{Advances in Neural Information Processing Systems}.

\bibitem[{Pennington et~al.(2014)Pennington, Socher, and Manning}]{glove}
Jeffrey Pennington, Richard Socher, and Christopher Manning. 2014.
\newblock \href {https://doi.org/10.3115/v1/D14-1162} {{G}lo{V}e: Global
  vectors for word representation}.
\newblock In \emph{Proceedings of the 2014 Conference on Empirical Methods in
  Natural Language Processing ({EMNLP})}, pages 1532--1543, Doha, Qatar.
  Association for Computational Linguistics.

\bibitem[{Qiang et~al.(2020{\natexlab{a}})Qiang, Li, Yi, Yuan, and
  Wu}]{qiang2020BERTLS}
Jipeng Qiang, Yun Li, Zhu Yi, Yunhao Yuan, and Xindong Wu. 2020{\natexlab{a}}.
\newblock Lexical simplification with pretrained encoders.
\newblock \emph{Thirty-Fourth AAAI Conference on Artificial Intelligence}, page
  8649–8656.

\bibitem[{Qiang et~al.(2020{\natexlab{b}})Qiang, Li, Zhu, Yuan, and
  Wu}]{DBLP:journals/corr/abs-2006-14939}
Jipeng Qiang, Yun Li, Yi~Zhu, Yunhao Yuan, and Xindong Wu. 2020{\natexlab{b}}.
\newblock \href {http://arxiv.org/abs/2006.14939} {{LSBert}: {A} simple
  framework for lexical simplification}.
\newblock \emph{CoRR}, abs/2006.14939.

\bibitem[{Qiang et~al.(2023)Qiang, Liu, Li, Yuan, and
  Zhu}]{qiang-etal-2023-parals}
Jipeng Qiang, Kang Liu, Yun Li, Yunhao Yuan, and Yi~Zhu. 2023.
\newblock \href {https://doi.org/10.18653/v1/2023.acl-long.206} {{P}ara{LS}:
  Lexical substitution via pretrained paraphraser}.
\newblock In \emph{Proceedings of the 61st Annual Meeting of the Association
  for Computational Linguistics (Volume 1: Long Papers)}, pages 3731--3746,
  Toronto, Canada. Association for Computational Linguistics.

\bibitem[{Raffel et~al.(2020)Raffel, Shazeer, Roberts, Lee, Narang, Matena,
  Zhou, Li, and Liu}]{t5}
Colin Raffel, Noam Shazeer, Adam Roberts, Katherine Lee, Sharan Narang, Michael
  Matena, Yanqi Zhou, Wei Li, and Peter~J. Liu. 2020.
\newblock \href {http://jmlr.org/papers/v21/20-074.html} {Exploring the limits
  of transfer learning with a unified text-to-text transformer}.
\newblock \emph{Journal of Machine Learning Research}, 21(140):1--67.

\bibitem[{Saggion et~al.(2022)Saggion, {\v{S}}tajner, Ferr{\'e}s, Sheang,
  Shardlow, North, and Zampieri}]{saggion-etal-2022-findings}
Horacio Saggion, Sanja {\v{S}}tajner, Daniel Ferr{\'e}s, Kim~Cheng Sheang,
  Matthew Shardlow, Kai North, and Marcos Zampieri. 2022.
\newblock \href {https://aclanthology.org/2022.tsar-1.31} {Findings of the
  {TSAR-2022} shared task on multilingual lexical simplification}.
\newblock In \emph{Proceedings of the Workshop on Text Simplification,
  Accessibility, and Readability (TSAR-2022)}, pages 271--283, Abu Dhabi,
  United Arab Emirates (Virtual). Association for Computational Linguistics.

\bibitem[{Sellam et~al.(2020)Sellam, Das, and Parikh}]{sellam-etal-2020-bleurt}
Thibault Sellam, Dipanjan Das, and Ankur Parikh. 2020.
\newblock \href {https://doi.org/10.18653/v1/2020.acl-main.704} {{BLEURT}:
  Learning robust metrics for text generation}.
\newblock In \emph{Proceedings of the 58th Annual Meeting of the Association
  for Computational Linguistics}, pages 7881--7892, Online. Association for
  Computational Linguistics.

\bibitem[{Souza et~al.(2020)Souza, Nogueira, and Lotufo}]{souza2020bertimbau}
F{\'a}bio Souza, Rodrigo Nogueira, and Roberto Lotufo. 2020.
\newblock {BERTimbau}: Pretrained {BERT} models for {Brazilian Portuguese}.
\newblock In \emph{Intelligent Systems}, pages 403--417, Cham. Springer
  International Publishing.

\bibitem[{Speer(2022)}]{robyn_speer_2022_7199437}
Robyn Speer. 2022.
\newblock \href {https://doi.org/10.5281/zenodo.7199437} {rspeer/wordfreq:
  v3.0}.
\newblock https://doi.org/10.5281/zenodo.7199437.

\bibitem[{Wada et~al.(2022)Wada, Baldwin, Matsumoto, and
  Lau}]{wada-etal-2022-unsupervised}
Takashi Wada, Timothy Baldwin, Yuji Matsumoto, and Jey~Han Lau. 2022.
\newblock \href {https://aclanthology.org/2022.coling-1.366} {Unsupervised
  lexical substitution with decontextualised embeddings}.
\newblock In \emph{Proceedings of the 29th International Conference on
  Computational Linguistics}, pages 4172--4185, Gyeongju, Republic of Korea.
  International Committee on Computational Linguistics.

\bibitem[{Wada et~al.(2023)Wada, Matsumoto, Baldwin, and
  Lau}]{wada-etal-2023-unsupervised}
Takashi Wada, Yuji Matsumoto, Timothy Baldwin, and Jey~Han Lau. 2023.
\newblock \href {https://doi.org/10.18653/v1/2023.findings-acl.290}
  {Unsupervised paraphrasing of multiword expressions}.
\newblock In \emph{Findings of the Association for Computational Linguistics:
  ACL 2023}, pages 4732--4746, Toronto, Canada. Association for Computational
  Linguistics.

\bibitem[{Whistely et~al.(2022)Whistely, Mathias, and
  Poornima}]{whistely-etal-2022-presiuniv}
Peniel Whistely, Sandeep Mathias, and Galiveeti Poornima. 2022.
\newblock \href {https://aclanthology.org/2022.tsar-1.22} {{P}resi{U}niv at
  {TSAR}-2022 shared task: Generation and ranking of simplification substitutes
  of complex words in multiple languages}.
\newblock In \emph{Proceedings of the Workshop on Text Simplification,
  Accessibility, and Readability (TSAR-2022)}, pages 213--217, Abu Dhabi,
  United Arab Emirates (Virtual). Association for Computational Linguistics.

\bibitem[{Xue et~al.(2021)Xue, Constant, Roberts, Kale, Al-Rfou, Siddhant,
  Barua, and Raffel}]{mt5}
Linting Xue, Noah Constant, Adam Roberts, Mihir Kale, Rami Al-Rfou, Aditya
  Siddhant, Aditya Barua, and Colin Raffel. 2021.
\newblock \href {https://doi.org/10.18653/v1/2021.naacl-main.41} {m{T}5: A
  massively multilingual pre-trained text-to-text transformer}.
\newblock In \emph{Proceedings of the 2021 Conference of the North American
  Chapter of the Association for Computational Linguistics: Human Language
  Technologies}, pages 483--498, Online. Association for Computational
  Linguistics.

\bibitem[{Štajner et~al.(2022)Štajner, Ferrés, Shardlow, North, Zampieri,
  and Saggion}]{10.3389/frai.2022.991242}
Sanja Štajner, Daniel Ferrés, Matthew Shardlow, Kai North, Marcos Zampieri,
  and Horacio Saggion. 2022.
\newblock \href {https://doi.org/10.3389/frai.2022.991242} {Lexical
  simplification benchmarks for {English}, {Portuguese}, and {Spanish}}.
\newblock \emph{Frontiers in Artificial Intelligence}, 5.

\end{thebibliography}
\bibliographystyle{acl_natbib}

\newpage

\appendix

\section{Impact of Clustering}\label{ablation_study_appendix}
\begin{table*}[t!]
\begin{center}
\begin{adjustbox}{max width=\textwidth}
\begin{tabular}{m{0.02\linewidth}m{0.18\linewidth}m{0.85\linewidth}}
\toprule
\multicolumn{2}{l}{ Context ($x$ = \ex{elite}) }&Syria is overwhelmingly Sunni, but President Bashar Assad and the ruling \underline{elite} belong to the minatory Alawite sect.\\\midrule
\multicolumn{2}{l}{Gold} &upper class, class, establishment, nobility, aristocracy, circle, elect, group, high society, noble, privileged, rich, select, society, superiors
\\\midrule
\multicolumn{2}{l}{\citet{wada-etal-2022-unsupervised}} &\textbf{establishment}, hierarchy, wealthy, bureaucracy, apparatus, leadership, ruling, affluent, clergy, mafia\\\midrule
\multirow{7}{*}{T5}&Cluster1 ($w_k=0$) &special, military, small, specialized, american, professional, secret, infamous, heroic, undercover\\
&Cluster2 ($w_k=5$)&\textbf{class}, political, \textbf{privileged}, \textbf{rich}, majority, minority, party, \textbf{group}, wealthy, liberal\\
&Cluster3 ($w_k=1$)&exclusive, international, prestigious, new, small, professional, top, large, \textbf{select}, special\\
&Cluster4 ($w_k=0$)&top, professional, great, competitive, good, high, olympic, pro, collegiate, excellent\\
\cmidrule{2-3}
&\multirow{1}{*}{Soft Retrieval} &\textbf{class}, \textbf{privileged}, political, \textbf{rich}, majority, minority, wealthy, exclusive, party, \textbf{group}\\
&\multirow{1}{*}{No Clustering} &top, professional, exclusive, international, special, great, prestigious, small, new, \textbf{select}\\\midrule
\multicolumn{2}{l}{OURS} &\textbf{class}, \textbf{establishment}, leadership, \textbf{rich}, hierarchy, \textbf{privileged}, bureaucracy, apparatus, family, clergy\\
\multicolumn{2}{l}{GPT-3.5-turbo-ens}&
ruling class, \textbf{high society}, \textbf{aristocracy}, \textbf{upper class}, exclusive, \textbf{nobility}, \textbf{privileged}, privileged few, \textbf{establishment}, top brass\\\midrule\midrule

\multicolumn{2}{l}{Context ($x$ = \ex{extend}) }&I would wish to \underline{extend} my thoughts and prayers to the family and friends of the victim at this terrible time.\\\midrule
\multicolumn{2}{l}{Gold} &expand, offer, give, send, continue, convey, dedicate, relay, reveal, share
\\\midrule

\multicolumn{2}{l}{\citet{wada-etal-2022-unsupervised}} &express, expanded, expanding, \textbf{offer}, render, impart, \textbf{convey}, exert, confer, grant\\\midrule
\multirow{6}{*}{T5}&Cluster1 ($w_k=1$)& reach, go, stretch, run, apply, \textbf{continue}, spread, move, amount, span\\
&Cluster2 ($w_k=0$)&increase, prolong, lengthen, improve, shorten, reduce, \textbf{continue}, stretch, change, make\\
&Cluster3 ($w_k=2$)&enhance, provide, apply, improve, increase, bring, \textbf{offer}, broaden, use, \textbf{give}\\
&Cluster4 ($w_k=4$)&\textbf{offer}, \textbf{give}, \textbf{send}, express, say, \textbf{convey}, wish, add, make, provide\\
\cmidrule{2-3}
&Soft Retrieval&\textbf{offer}, \textbf{give}, \textbf{send}, provide, express, apply, enhance, make, add, bring\\
&No Clustering&increase, improve, stretch, prolong, \textbf{give}, enhance, \textbf{continue}, \textbf{offer}, apply, provide\\\midrule
\multicolumn{2}{l}{OURS}&\textbf{offer}, express, \textbf{send}, spread, stretch, \textbf{convey}, \textbf{give}, \textbf{share}, lend, extensions\\
\multicolumn{2}{l}{GPT-3.5-turbo-ens}&\textbf{offer}, \textbf{send}, express, \textbf{convey}, \textbf{share}, stretch, lengthen, \textbf{give}, present, \textbf{expand}\\

\bottomrule
\end{tabular}
\end{adjustbox}
\end{center}
\caption{Examples of top-10 substitutes generated based on the target context (\citet{wada-etal-2022-unsupervised}; \secref{generation_tgt}) and the augmented contexts (\secref{generation_sample}). ``OURS'' denotes the substitutes obtained by reranking the candidates of \citet{wada-etal-2022-unsupervised} and ``Soft Retrieval''. The values for $w_k$ denote the weights for each cluster in \eqnref{t5_score}, which correspond to the number of the shared words between the top-15 words from \citet{wada-etal-2022-unsupervised} and the top-25 words from each cluster. The words included in the gold labels are boldfaced. }\label{examples}
\end{table*}

\tabref[]{examples} shows two examples where clustering plays a crucial role; the first instance was partially shown in \tabref{ablation_example} and discussed in \secref{ablation_study}. In the second instance, Soft Retrieval retrieves substitutes that are relevant to the meaning of the target word \ex{extend} given this particular context. Without clustering, on the other hand, we get a mixed bag of words that represent different senses of \ex{extend} (with more frequent senses ranked higher). In both cases, GPT-3.5 produces context-aware substitutes, although this is not always the case (as we discuss in the next section), and some of the candidates do not fit naturally in context (e.g.\ \ex{ruling class} is predicted as the best substitute for \ex{elite} used in \ex{{ruling} {elite}}).

\section{Error Analysis}

\tabref[]{error_analysis_example} shows some examples of outputs from our model and GPT-3.5-turbo-ens on the lexical simplification and substitution tasks. In the first example, our model generates near-synonyms of the target word \ex{infamous} with both negative and positive connotative meaning (e.g.\ \ex{notorious} and \ex{famed}, respectively), while GPT-3.5 generates negative-connotation words like \ex{disgraceful} only. This is because our model heavily relies on word representations and assigns high scores to those words that appear in similar contexts. Similarly, our model incorrectly predicts \ex{notoriously} as a substitute for \ex{infamous} despite its ungrammaticality in this context, likely due to the similarity of their embeddings; we surmise that using a part-of-speech tagger would help alleviate this problem.\footnote{Another idea is to increase the weight for the LM perplexity in reranking, but it comes with a trade-off as it increases the unwanted bias of the target context.} In the second instance, our model is overly affected by the target context and generates words that often appear in similar contexts to the target context but have different semantics from the target word \ex{strategic}, e.g.\ \ex{global} and \ex{economic}. In comparison, GPT-3.5 generates more correct substitutes; however, some of them do not sound quite natural in this context (e.g.\ \ex{calculated}). This is more evident in the third instance, where some of the ``gold'' substitutes are semantically marked when put into this context --- the original phrase \ex{in defiance of calls} means ``opposition against calls'', but \ex{in resistance of calls} and \ex{in rebellion of calls} (both predicted by GPT-3.5 and included in the gold labels) do not sound natural. These examples suggest that human annotators are sometimes oblivious to the context and consider substitutes largely based on the out-of-context similarities of the words,\footnote{Similarly, in Spanish and Portuguese, annotators suggest both masculine and feminine nouns given a nominal target word. \citet{wada-etal-2022-unsupervised} make a similar observation in Italian.} which motivates a call for revisiting the exact goal of lexical simplification/substitution and its annotation schemes, e.g.\ whether the words should be annotated based on the {\it similarity of lexical semantics} or {\it acceptability in context}.  The same concern is also raised by the strong performance of GPT-3.5 without access to the target context (\tabref[]{result_tsar}).

In the last three examples, which are taken from the SWORDS lexical substitution data set, the sensitivity of our model to context works favourably and results in better substitutes than GPT-3.5, which, in those examples, generates substitutes without considering the context very much (in contrast to the examples in \tabref{error_analysis_example}). For these instances, we also tried using the ChatGPT web interface (the free version, accessed in May 2023) and found that its outputs are highly stochastic even with the same prompt:\footnote{We have also found that it often outputs an opening line such as ``Here are ten alternative words for ...'', while GPT-3.5-turbo (accessed via OpenAI API) usually returns a list of substitutes only.} sometimes it returns substitutes that are quite similar to the ones generated by GPT-3.5-turbo, and other times it generates more context-aware and accurate substitutes (e.g.\ \ex{business} for \ex{service} and \ex{probably/presumably} for \ex{likely}). As such, further investigation is needed to see how carefully the model pays attention to the context (given different prompts), and how well it works for instances that require a profound understanding of the context. 

\label{error_analysis}

\begin{table*}[t!]
\begin{center}
\begin{adjustbox}{max width=\textwidth}
\begin{tabular}{m{0.15\linewidth}m{0.9\linewidth}}
\toprule

\makecell[l]{Context \\($x$ = \ex{infamous}) }& He has denied charges of genocide, murder acts of terror and other crimes against humanity.The most \underline{infamous} of the charges accuses Mladic of overseeing the massacre  of 8 thousand Muslim boys and men in Srebrenica in Eastern Bosnia in 1995.
\\\midrule
OURS &\textbf{notorious}, iconic, famed, renowned, legendary, \textbf{controversial}, heinous, notoriously, dreaded, notoriety
\\\midrule
GPT-3.5&\textbf{notorious}, \textbf{disreputable}, scandalous, shameful, ill-famed, infamously known, \textbf{disgraceful}, \textbf{dishonorable}, ignominious, detestable
\\\midrule\midrule

\makecell[l]{Context \\($x$ = \ex{strategic}) }& The Taliban said it was in response to Obama's visit and to the \underline{strategic} partnership deal he signed with Afghan President Hamid Karzai, a pact that sets out a long-term U.S. role after most foreign combat troops leave by the end of 2014.
\\\midrule
OURS &
security, \textbf{political}, national, international, global, military, economic, \textbf{tactical}, operational, policy
\\\midrule
GPT-3.5&\textbf{important}, \textbf{crucial}, essential, vital, significant, \textbf{planned}, key, critical, \textbf{tactical}, \textbf{calculated}\\\midrule\midrule

\makecell[l]{Context \\($x$ = \ex{defiance}) }&  Taliban bombers attacked a heavily fortified guesthouse used by Westerners in Kabul on Wednesday, announcing the start of their annual in \underline{defiance} of calls from visiting US President Barack Obama that the war was ending.
\\\midrule
OURS & \textbf{disregard}, violation, contravention, contempt, \textbf{spite}, rejection, breach, defying, \textbf{opposition}, contradiction
\\\midrule
GPT-3.5&\textbf{resistance}, \textbf{rebellion}, \textbf{opposition}, disobedience, challenge, noncompliance, insubordination, \textbf{refusal}, nonconformity, dissent\\
\midrule\midrule
\makecell[l]{Context \\($x$ = \ex{form}) }& "Perhaps." "I think his face is his \underline{form} of vanity. It's the reverse of you with those ridiculous stomach muscles.
\\\midrule
OURS &\textbf{type}, mode, version, method, manner, means, style, way, kind, breed
\\\midrule
GPT-3.5 &shape, figure, structure, configuration, physique, appearance, build, outline, contour, profile
\\
\midrule\midrule
\makecell[l]{Context \\($x$ = \ex{service})} & The National Association of Diaper Services, Philadelphia, says that since January it has gotten more than 672 inquiries from people interested in starting diaper services. Elisa Hollis launched a diaper \underline{service} last year because State College, Pa., where she lives, didn't have one. Diaper shortages this summer limited growth at Stork Diaper Services, Springfield, Mass., where business is up 25\% in
\\\midrule
OURS &\textbf{business}, \textbf{company}, provider, program, center, operation, system, delivery, site, agency
\\\midrule
GPT-3.5 &assistance, support, aid, help, provision, maintenance, care, repair, supply, benefit
\\

\midrule\midrule

\makecell[l]{Context \\($x$ = \ex{likely})} & Upjohn further estimated that about 50\% of the employees who leave for early retirement may be replaced. As a result, Upjohn will \underline{likely} trim only about 275 to 350 of its more than 21,000 jobs world-wide. In composite trading on the New York Stock Exchange yesterday, Upjohn shares rose 87.5 cents to \$38.875 apiece.
\\\midrule
OURS &\textbf{probably}, undoubtedly, probable, possibly, surely, certainly, perhaps, \textbf{presumably}, potentially, likelihood
\\\midrule
GPT-3.5 &probable, expected, anticipated, possible, plausible, foreseeable, credible, presumed, expect, anticipate
\\

\bottomrule
\end{tabular}
\end{adjustbox}
\end{center}
\caption{Examples of outputs from our model and GPT-3.5-turbo-ens. The first and last three instances are from lexical simplification and substitution data sets, respectively. The words included in the gold labels are boldfaced.}\label{error_analysis_example}
\end{table*}

\begin{table*}[t!]
\begin{center}
\begin{adjustbox}{max width=\textwidth}
\begin{tabular}{llclllllllllllllllllllllll}
\toprule
&\multirow{2}{*}{Model}
&\multirow{2}{*}{\textbf{ACC@1}}
&&\multicolumn{3}{c}{\textbf{ACC@k@Top1}}
&&\multicolumn{3}{c}{\textbf{MAP@k}}
&&\multicolumn{3}{c}{{\textbf{Potential@k}}}\\
  \cmidrule{5-7}
  \cmidrule{9-11}
  \cmidrule{13-15}
&
&
&&\multicolumn{1}{c}{\textbf{k=1}}&\multicolumn{1}{c}{\textbf{k=2}}&\multicolumn{1}{c}{\textbf{k=3}}
&&\multicolumn{1}{c}{\textbf{k=3}}&\multicolumn{1}{c}{\textbf{k=5}}&\multicolumn{1}{c}{\textbf{k=10}}
&&\multicolumn{1}{c}{\textbf{k=3}}&\multicolumn{1}{c}{\textbf{k=5}}&\multicolumn{1}{c}{\textbf{k=10}}
\\

  \midrule
\multirow{13}{*}{en}&GPT-3.5-text-davinci-002-zero
&77.2&&42.6&53.4&57.1&&50.9&36.5&20.9&&89.0&93.0&94.4
\\
&GPT-3.5-text-davinci-002-ens
&81.0&&42.9&61.1&68.6&&58.3&44.9&28.1&&96.2&98.1&99.5\\

&GPT-3.5-turbo-zero
&82.8&&52.3&63.5&68.1&&60.9&46.7&28.0&&92.8&94.4&95.2
\\

&GPT-3.5-turbo-zero (w/o context)
&83.9&&46.6&60.9&67.6&&62.4&46.5&28.3&&92.8&95.4&97.3
\\

&GPT-3.5-turbo-ens
&87.4&&54.7&65.7&71.8&&65.5&52.5&33.3&&97.3&99.2&\textbf{100.0}\\

&GPT-3.5-turbo-ens (w/o context)
&82.6&&46.1&63.8&70.8&&64.8&50.0&31.9&&94.6&97.6&98.9\\

&MANTIS
&65.7&&31.9&45.0&53.9&&47.3&36.0&21.9&&87.7&94.6&97.9\\
&LSBert 
&59.8&&30.3&44.5&53.1&&40.8&29.6&17.6&&82.3&87.7&94.6\\
&\citet{wada-etal-2022-unsupervised} + fastText
&64.6&&27.6&44.2&51.2&&43.7&32.2&19.8&&86.9&91.4&95.2\\

&\textbf{OURS} (BERT + mT5)
&69.4&&33.8&49.1&58.7&&45.0&34.2&21.0&&88.2&94.9&98.1\\

&\textbf{OURS}  (DeBERTa-V3 + T5)
&79.9&&43.7&57.9&63.5&&57.5&42.9&26.6&&94.1&97.6&98.9\\\midrule
&WordFreq + GPT-3.5-turbo-ens
&\textbf{89.3}&&55.0&\textbf{68.1}&73.2&&68.2&54.1&34.6&&97.9&\textbf{99.7}&99.7\\

& + \textbf{OURS}
&\textbf{89.3}&&\textbf{56.0}&\textbf{68.1}&\textbf{75.1}&&\textbf{69.9}&\textbf{55.2}&\textbf{35.4}&&\textbf{98.7}&\textbf{99.7}&\textbf{100.0}\\

\midrule\midrule

\multirow{12}{*}{pt}&GPT-3.5-text-davinci-002-zero
&63.6&&37.2&46.3&51.6&&41.0&28.9&16.2&&78.6&81.8&84.2\\
&GPT-3.5-text-davinci-002-ens
&77.0&&43.6&53.5&62.3&&50.1&36.2&21.7&&91.7&94.9&97.9\\

&GPT-3.5-turbo-zero
&79.4&&46.5&59.1&63.6&&51.0&37.1&21.7&&88.8&89.8&91.4
\\

&GPT-3.5-turbo-zero (w/o context)
&78.6&&44.1&57.2&62.8&&51.4&38.0&22.1&&90.6&92.8&94.4\\

& GPT-3.5-turbo-ens
&85.8&&48.7&65.2&73.5&&58.7&44.5&26.8&&\textbf{97.3}&98.4&98.9
\\

&GPT-3.5-turbo-ens (w/o context)
&84.5&&\textbf{49.5}&64.4&69.8&&57.3&43.2&25.8&&94.9&96.5&97.6\\

&GMU-WLV
&48.1&&25.4&37.2&39.6&&28.2&19.7&11.5&&68.7&75.7&84.0\\
&LSBert 
&32.6&&15.8&23.3&28.6&&19.0&13.1&7.8&&49.5&58.0&67.4\\
&\citet{wada-etal-2022-unsupervised} + fastText
&51.9&&27.0&37.7&42.8&&31.0&22.5&12.9&&70.9&77.3&85.3\\
&\textbf{OURS} (BERT + mT5)
&61.5&&31.6&45.2&52.7&&38.0&28.3&16.9&&83.2&89.3&92.8
\\\midrule
&WordFreq + GPT-3.5-turbo-ens
&85.6&&48.4&65.5&74.9&&59.9&45.4&27.4&&\textbf{97.3}&98.1&\textbf{99.2}\\

& + \textbf{OURS}
&\textbf{86.4}&&49.2&\textbf{66.8}&\textbf{76.7}&&\textbf{61.1}&\textbf{46.8}&\textbf{28.1}&&\textbf{97.3}&\textbf{98.7}&\textbf{99.2}\\

\midrule\midrule
\multirow{12}{*}{es}&GPT-3.5-text-davinci-002-zero
&57.1&&30.7&39.7&45.1&&35.3&24.5&13.8&&69.0&71.5&74.5
\\
&GPT-3.5-text-davinci-002-ens
&65.2&&35.1&51.1&57.9&&42.8&32.4&19.7&&82.1&88.9&94.0\\
&GPT-3.5-turbo-zero
&64.4&&33.4&43.8&50.5&&45.2&33.0&18.8&&75.0&77.4&78.0
\\

& GPT-3.5-turbo-zero (w/o context)
&74.7&&40.2&50.3&57.1&&52.5&37.9&22.0&&87.2&88.6&89.9\\

& GPT-3.5-turbo-ens
&76.4&&39.9&54.3&62.2&&55.9&42.2&25.7&&89.1&92.4&94.6
\\
& GPT-3.5-turbo-ens (w/o context)
&76.1&&\textbf{42.4}&52.7&59.5&&54.1&41.3&25.0&&88.3&91.8&94.6
\\

&PresiUniv 
&37.0&&20.4&27.7&32.9&&21.5&15.0&8.3&&58.4&64.7&72.6\\
&LSBert 
&28.8&&9.5&14.4&18.2&&18.7&13.5&8.0&&49.5&61.1&74.7
\\
&\citet{wada-etal-2022-unsupervised} + fastText
&32.3&&15.5&21.2&25.0&&19.5&13.5&7.8&&49.2&53.5&62.8\\

&\textbf{OURS} (BERT + mT5)
&47.8&&22.0&31.8&37.0&&30.0&21.5&13.0&&71.5&80.7&88.3

\\\midrule
&WordFreq + GPT-3.5-turbo-ens
&\textbf{78.3}&&42.1&54.1&65.2&&57.6&43.9&26.6&&89.4&92.9&\textbf{96.2}\\

&+ \textbf{OURS}
&77.7&&41.8&\textbf{56.5}&\textbf{66.8}&&\textbf{59.1}&\textbf{44.8}&\textbf{27.2}&&\textbf{89.9}&\textbf{93.2}&\textbf{96.2}\\

\bottomrule
\end{tabular}
\end{adjustbox}
\end{center}
\caption{The results on the TSAR-2022 lexical simplification task. ``-zero/ens'' denote the zero-shot/ensemble models proposed by UniHD \cite{aumiller-gertz-2022-unihd}, and ``w/o context'' indicates the performance without access to the target context. The best scores in each language are boldfaced.}
\label{result_tsar_all}
\end{table*}

\begin{table*}[t]
\begin{center}
\begin{adjustbox}{max width=\textwidth}
\begin{tabular}{m{1.0\linewidth}}
\toprule
        \multicolumn{1}{c}{GPT-3.5-turbo Prompt With Context} 
        \\\midrule
        \{"role": "system", "content": "You are a helpful assistant."\}\\
        \{"role": "user", "content": "Context: A local witness said a separate group of attackers disguised in burqas — the head-to-toe robes worn by conservative Afghan women — then tried to storm the compound.\textbackslash nQuestion: Given the above context, list ten alternative words for "disguised" that are easier to understand.\textbackslash n"\}\\
        \{"role": "assistant","content": "1. concealed\textbackslash n2. dressed\textbackslash n3. hidden\textbackslash n4. camouflaged\textbackslash n5. changed\textbackslash n6. covered\textbackslash n7. masked\textbackslash n8. unrecognizable\textbackslash n9 converted\textbackslash n10. impersonated\textbackslash n\textbackslash n"\}\\
        \{"role": "user", "content": "Context: \{CONTEXT\}\textbackslash nQuestion: Given the above context, list ten alternative words for "{WORD}" that are easier to understand.\textbackslash n"\}\\\midrule
        \multicolumn{1}{c}{GPT-3.5-turbo Prompt Without Context} \\\midrule
        
        \{"role": "system", "content": "You are a helpful assistant."\}\\
        \{"role": "user", "content": "Question: Find ten easier words for "compulsory".\textbackslash n"\}\\
        \{"role": "assistant", "content": "1. mandatory\textbackslash n2. required\textbackslash n3. essential\textbackslash n4. forced\textbackslash n5. important\textbackslash n6. necessary\textbackslash n7. obligatory\textbackslash n8. unavoidable\textbackslash n9. binding\textbackslash n10. prescribed\textbackslash n\textbackslash n"\}\\
        \{"role": "user", "content": "Question: Find ten easier words for "{WORD}"".\textbackslash n\}
\\
\bottomrule
\end{tabular}
\end{adjustbox}
\caption{The prompt template for one-shot GPT-3.5-turbo in English with and without context. WORD and CONTEXT denote the target word $x$ and context $c_x$, respectively. We modified the template used in \citet{aumiller-gertz-2022-unihd} for the purpose of using \ex{gpt-3.5-turbo} instead of \ex{text-davinici-002}.}\label{prompt_template}
\end{center}
\end{table*}

\end{document}